\documentclass[11pt]{article}
\usepackage[margin=1in]{geometry}
\usepackage{graphicx}
\usepackage{booktabs}
\usepackage{amsmath,amssymb}
\usepackage[colorlinks=true,linkcolor=blue,citecolor=blue,urlcolor=blue]{hyperref}
\usepackage{caption}
\title{\textbf{Shared-Prefix KV Reuse Across Standard LoRA Adapters:\\ Quality and Serving Tradeoffs}}
\author{
  Dushyant Rajput \\[2pt]
  AltSlate Labs LLP \\
  \texttt{dushyant@altslate.com}
}
\date{September 2026}

\begin{document}
\maketitle

\begin{abstract}
A common small-model deployment runs one shared backbone with several LoRA~\cite{lora} specialists that
answer over the same context. Serving them na\"ively re-prefills that shared context once per specialist.
We study a narrow, practical question: for \textbf{already-trained standard} LoRA adapters---not adapters
retrained for cache compatibility---how much task quality is preserved if the backbone's prefill KV cache
is computed once and reused across specialists, and what does that buy in serving cost? On a Qwen3-1.7B
\cite{qwen3} backbone with two adapters (extractive QA on HotpotQA~\cite{hotpotqa}, arithmetic reasoning on
GSM8K~\cite{gsm8k}), we sweep the boundary at which the specialist takes over from the reused base cache and
measure paired quality differences and serving cost. \textbf{Full-prefix reuse had the lowest prefill cost
and a small quality difference on held-out GSM8K ($\Delta = -4.6$ EM at a 160-token budget; $-3.0$ at 320
tokens; $-0.8$ under a second training seed---all favoring native, only the first excluding zero, and the
magnitude not consistent). Partial recomputation provided no demonstrated advantage. Neither quality
equivalence nor a general boundary-selection rule is established.} We also report a closed-form ridge KV
translator that did not beat direct reuse, and specialist-dependence contrasts whose intervals all include
zero. The measured serving benefit is \textbf{warm-cache time-to-first-token}, which grows with context
(${\approx}16\times$ at 8K); two-branch peak memory was only 12\% lower and, on inspection, the prefix was
\emph{never physically shared} across branches---this implementation reuses KV \emph{values} but copies
their storage, so shared-cache memory savings are not achieved.
\end{abstract}

\section{Introduction}
Small models are increasingly deployed as a composable system: one backbone plus several lightweight
LoRA~\cite{lora} specialists, routed per request. These specialists often answer over a \emph{shared}
context---the same retrieved passages, the same conversation history, the same few-shot demonstrations---and
differ only in the adapter applied. Serving such a system the obvious way re-runs the prefill over that
shared context once per specialist that touches it.

Prefix caching in serving stacks (PagedAttention~\cite{pagedattention}, RadixAttention~\cite{sglang})
already avoids recomputation when the \emph{same} model sees an \emph{identical} prefix. The question here
is different: several \emph{different} models (backbone $+\,A_i$) see the \emph{same} prefix. For a family of
\emph{different-size} models this needs a learned map between key/value spaces---NVIDIA's closed-form linear
KV transfer~\cite{crossmodelkv} and the concurrent CacheBridge~\cite{cachebridge}. Activated
LoRA~\cite{alora} and its serving engine~\cite{aloraserving} instead modify the adapter so a base prefix
cache is exactly reusable by construction.

We study the case those methods do not target: \textbf{standard} LoRA adapters, already trained for their
task with no cache-compatibility objective, on a \textbf{shared backbone}. The precise question is:

\begin{quote}
\emph{How much task quality does inference-time base-cache reuse preserve for already-trained standard LoRA
adapters, without retraining for cache compatibility, and at what serving cost?}
\end{quote}

Our answer is qualified. We report the exact reuse procedure, a quality study with paired confidence
intervals, a reconciled serving-cost accounting, one rejected translator, and specialist-dependence
contrasts that do not resolve. We are explicit about which measurements are exploratory and which we would
trust, and about the gap between the context lengths used for quality and for serving cost.

\section{Setting and method}
\paragraph{Composable serving.} One backbone $B$ (Qwen3-1.7B~\cite{qwen3}) hosts LoRA specialists, each
rank-16 ($\alpha=32$) on the attention projections (q/k/v/o). We train two: QA on HotpotQA~\cite{hotpotqa}
and math on GSM8K~\cite{gsm8k} (recipe in Appendix~\ref{app-repro}). At request time a router selects a
specialist; several specialists may answer over the same shared context. Because every specialist shares the
backbone weights (loaded once) and the same context, the shared prefix is a candidate for compute-once
reuse. Throughout, $M$ denotes the number of specialists sharing a context; the two trained adapters are
used for quality, and $M$ enters only the (adapter-agnostic) serving-cost accounting of \S\ref{sec-cost}.

\begin{figure}[h]\centering
\includegraphics[width=0.96\textwidth]{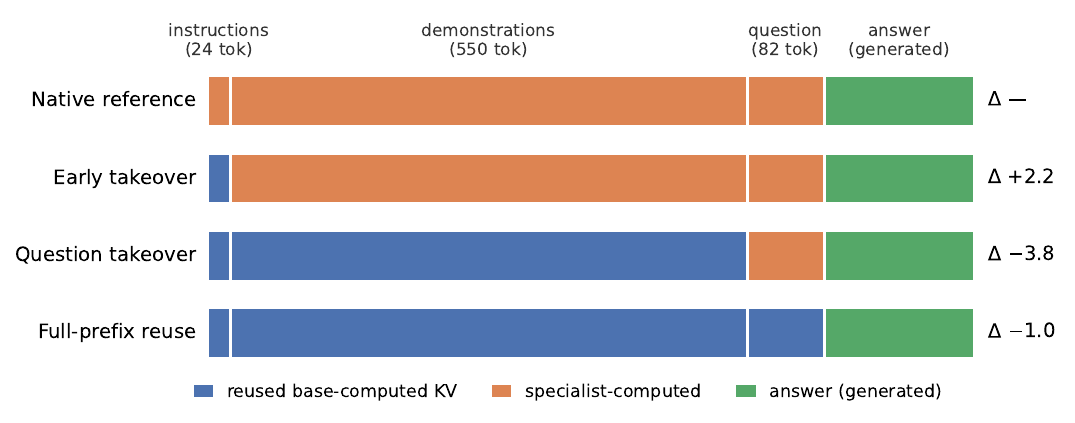}
\caption{The four takeover boundaries. Blue tokens reuse the base-computed prefill KV; orange tokens are
processed by the specialist; green is the generated answer. In \texttt{full-prefix} the specialist still
processes the final prompt token (a one-token orange sliver, not drawn to scale). The full prompt and
decoding are identical across rows; only the boundary $b$ moves.}
\label{fig-schematic}
\end{figure}

\paragraph{Reuse procedure.} The backbone computes the prefill KV cache for the shared prefix \emph{with
adapters disabled} (base representation). A chosen specialist then processes the \emph{non-reused suffix} of
the prompt on top of that cache and generates. No key/value mapping is applied. Crucially, the first answer
token's logits always come from the \emph{specialist}, which processes at least one prompt token
(Figure~\ref{fig-schematic}, Figure~\ref{lst-reuse}). For a boundary $b$ (number of reused prefix tokens)
and prompt of length $T$:

\begin{figure}[h]\centering
\begin{minipage}{0.95\textwidth}\scriptsize
\begin{verbatim}
prefix_kv = forward(B, tokens[0:b], adapters_disabled).kv   # b reused base-computed tokens
set_adapter(specialist)                                      # base -> specialist
out = forward(specialist, tokens[b:T], past=prefix_kv)       # suffix; positions continue at b..T-1
logits_1 = out.logits[-1]                                    # FIRST answer token: specialist's logits
# then greedy-decode with the specialist and the growing cache
\end{verbatim}
\end{minipage}
\caption{Reuse procedure. \texttt{native}: $b=0$ (specialist computes the whole prompt).
\texttt{early}: $b=|\text{instr}|$. \texttt{question}: $b=|\text{instr}|+|\text{demos}|$.
\texttt{full-prefix}: $b=T-1$, so the specialist processes exactly the final prompt token before
generating---not ``generation only.'' Position IDs continue from $b$; the reused cache is base-computed, the
suffix and all generation are specialist-computed.}
\label{lst-reuse}
\end{figure}

\paragraph{Takeover boundary.} We formalize the prompt as three segments---\texttt{[instructions |
demonstrations | question] -> answer}---and sweep $b$ across the four settings above
(Figure~\ref{fig-schematic}), holding the full prompt and greedy decoding fixed. This separates ``how many
tokens are reused'' from ``which content the specialist re-encodes.''

\paragraph{Evaluation.} Metrics are deterministic: token-level F1 for extractive QA, exact-match (final
numeric answer) for GSM8K; no model-based judging. We report paired bootstrap~\cite{bootstrap} 95\%
confidence intervals over per-example score differences (10,000 resamples), the appropriate interval for the
same-example, cached-vs-native comparisons here. Runs use transformers 5.5 / peft 0.20 / torch 2.11, bf16,
greedy decoding, generation capped at 160 tokens, on NVIDIA Blackwell GPUs. Exact revisions, prompts, sample
ranges, and timing methodology are in Appendix~\ref{app-repro}.

\section{Serving cost}\label{sec-cost}
We separate three quantities and, where possible, replace structural arithmetic with direct measurement. The
central distinction is between \textbf{logical} reuse (a specialist reuses previously-computed KV
\emph{values}, skipping their recomputation) and \textbf{physical} sharing (those values occupy one copy of
storage across branches). This implementation achieves the former; \S\ref{sec-nosharing} shows it does
\emph{not} achieve the latter.

\paragraph{Prefill count (structural).} When $M$ specialists answer over one shared context, the shared
prefix is prefilled once instead of $M$ times. Reuse does not make per-specialist work vanish: each
specialist still runs a suffix forward over its non-reused tokens plus generation
(Figure~\ref{lst-reuse}). Table~\ref{tab-cost} gives single-prefix cost; the native per-specialist prefill
(489 ms at 8K) exceeds the base prefill (413 ms) that reuse pays once.

\subsection{Warm-cache latency (measured)}
Table~\ref{tab-serving} reports latency across the three matched settings. TTFT for reuse is measured
\emph{warm-cache}: it excludes constructing the base prefix cache (that one-time cost---40/91/417 ms at
655/2K/8K---is amortized across specialists and reported separately) and \emph{includes} the specialist's
final-prompt-token forward (Figure~\ref{lst-reuse}). Warm-cache TTFT is where reuse wins, and the win grows
with context: at 8K, 30 ms vs.\ 486 ms, a \textbf{${\approx}16\times$ warm-cache TTFT speedup}. Because reuse
also generates \emph{longer} outputs on GSM8K (\S\ref{sec-quality}), a TTFT improvement does \emph{not} imply
a completion-latency improvement: completion was faster at 8K QA (265 vs.\ 582 ms) but slower on GSM8K (3149
vs.\ 2787 ms).

\begin{table}[h]\centering
\begin{tabular}{lrrrrrr}
\toprule
\textbf{setting} & \textbf{TTFT nat} & \textbf{TTFT reuse} & \textbf{compl nat} & \textbf{compl reuse} & \textbf{1-req peak nat} & \textbf{reuse} \\
\midrule
GSM8K ${\approx}655$ & 38 ms & 30 ms & 2787 ms & 3149 ms & 3.83 GB & 3.65 GB \\
QA 2K & 96 ms & 30 ms & 188 ms & 196 ms & 4.57 GB & 3.95 GB \\
QA 8K & 486 ms & 30 ms & 582 ms & 265 ms & 7.79 GB & 5.35 GB \\
\bottomrule
\end{tabular}
\caption{Warm-cache latency and single-request peak memory (means; GSM8K $n=500$, QA-2K $n=300$, QA-8K
$n=200$). Reuse TTFT excludes the one-time base prefill (40/91/417 ms) and includes the specialist's
final-token forward. Single-request peak is lower for reuse at long context because it skips native's
full-context prefill activation spike; this is a per-request working-set effect, not cross-branch sharing
(\S\ref{sec-nosharing}).}
\label{tab-serving}
\end{table}

\subsection{Two-branch memory: logical reuse, no physical sharing}\label{sec-nosharing}
We held two branches (QA + math) over one shared context and measured peak allocation across generation
(Table~\ref{tab-residency}). Two-branch peak was \textbf{12\% lower at 8K and 5\% lower at 2K} under reuse.
\textbf{This is not a sharing effect}: inspecting tensor storage, the two branches' prefix KV was never
physically shared---0\% of trials aliased the prefix, \emph{before or after} generation---because the cache
concatenates new keys/values each step, copying the prefix into each branch. The modest reduction comes from
reuse doing one base prefill instead of two adapter prefills, not from one copy of the prefix serving both
branches. Persistent shared-cache storage would require an implementation that preserves shared storage
during generation (a paged cache is one route); it \textbf{remains unimplemented here}.

\begin{table}[h]\centering
\begin{tabular}{rrrrc}
\toprule
\textbf{context} & \textbf{peak nat (2 br.)} & \textbf{peak reuse} & \textbf{reuse/native} & \textbf{prefix physically shared?} \\
\midrule
2048 & 4.62 GB & 4.38 GB & $0.95\times$ & no (0\%) \\
8192 & 7.92 GB & 6.98 GB & $0.88\times$ & no (0\%) \\
\bottomrule
\end{tabular}
\caption{Two simultaneously-retained branches (QA + math) over one shared context, $n=20$, 32 generated
tokens each. Reuse's peak is modestly lower (one base prefill vs.\ two), but the prefix is never physically
shared across branches---the saving is not from sharing.}
\label{tab-residency}
\end{table}

\begin{table}[h]\centering
\begin{tabular}{rrrr}
\toprule
\textbf{context (tok)} & \textbf{prefix KV cache} & \textbf{base prefill} & \textbf{specialist prefill} \\
\midrule
700 & 80 MB & 38 ms & 42 ms \\
2048 & 235 MB & 88 ms & 100 ms \\
8192 & 940 MB & 413 ms & 489 ms \\
\bottomrule
\end{tabular}
\caption{Single-prefix cost (Qwen3-1.7B, bf16; one benchmark run). Cache is the KV tensor footprint reuse
avoids recomputing; prefill is what reuse avoids repaying per specialist. Specialist prefill exceeds base by
the adapter's matrix-multiply overhead.}
\label{tab-cost}
\end{table}

\paragraph{Scope.} Backbone weights (3.5 GB) are resident once regardless of reuse; per-specialist
answer-side caches and generation buffers are unchanged. The serving benefit is warm-cache latency (largest
at long context), not a peak-memory reduction from sharing. The quality study (\S\ref{sec-quality}) uses
${\approx}655$-token GSM8K prompts and 2K/8K QA contexts, so quality and latency are reported at matched
scales; equivalence at any scale is not claimed (\S\ref{sec-limits}).

\section{Quality of reused-KV inference}\label{sec-quality}
\paragraph{Own-KV control (harness check).} Reusing each specialist's \emph{own} recomputed prefix KV should
be near-identical to native. It was (math EM 49.2 vs.\ 48.3, $n=120$): 3/120 examples disagreed (two
cached-correct, one native-correct). We did \emph{not} instrument the numerical divergence (e.g.\ per-layer
KV or first-token logit deltas), so we report the discrepancy as consistent with greedy sensitivity to
cache-reconstruction differences but \emph{not further diagnosed}. It is small relative to the cross-source
effects below.

\paragraph{Central result.} On 500 untouched GSM8K examples, full-prefix base-KV reuse reduced accuracy from
54.4\% to 49.8\% ($\Delta = -4.6$ percentage points; paired CI $[-8.8, -0.4]$; the point estimate moves to
$-3.0$ at a larger budget and $-0.8$ under a second seed---see \emph{Generation budget} and
\emph{Replication} below). In the 8K supplied-context QA workload, reuse reduced warm-cache TTFT from 486 ms
to 30 ms. Two-branch peak memory was 12\% lower, but storage inspection found no physical prefix sharing.
These results establish a quality--TTFT tradeoff; persistent shared-cache storage remains unimplemented.

\paragraph{Matched quality with a base-only baseline.} Table~\ref{tab-quality} gives the three conditions on
identical examples per setting. The adapter is necessary: base-only trails native by 46 EM on GSM8K and
27--29 F1 on QA---though, since base-only reaches the 160-token generation limit on 100\% of GSM8K examples,
this establishes adapter necessity \emph{under the tested decoding budget}, not budget-independent
inferiority. Full-prefix reuse costs a small but mostly significant amount of quality: $-4.6$ EM on GSM8K and
$-6.6$/$-4.5$ F1 on QA at 2K/8K.

\begin{table}[h]\centering
\begin{tabular}{lrrrrr}
\toprule
\textbf{setting (metric)} & \textbf{base-only} & \textbf{native} & \textbf{reuse} & \textbf{$\Delta$ reuse$-$native} & \textbf{cap\% nat/reuse} \\
\midrule
GSM8K, EM ($n{=}500$) & 8.4 & 54.4 & 49.8 & $-4.6\;[-8.8,-0.4]$ & 15 / 30 \\
QA 2K, F1 ($n{=}300$) & 42.7 & 69.4 & 62.8 & $-6.6\;[-10.5,-2.7]$ & 0 / 0.7 \\
QA 8K, F1 ($n{=}200$) & 43.3 & 72.5 & 67.9 & $-4.5\;[-9.3,+0.1]$ & 0 / 3 \\
\bottomrule
\end{tabular}
\caption{Matched quality on identical examples per setting (paired bootstrap CIs). base-only $\Delta$ vs
native is $-46.0$/$-26.6$/$-29.2$ (all excluding zero). GSM8K uses untouched \texttt{test[580:1080]}; QA uses
constructed 2K/8K contexts (gold paragraphs preserved, Appendix~\ref{app-repro}). cap\% is the fraction
reaching the 160-token limit.}
\label{tab-quality}
\end{table}

\paragraph{Generation budget.} Reuse reached the 160-token cap more often than native (30\% vs.\ 15\%). At a
pre-frozen 320-token budget on the same held-out examples, both conditions improve and cap-hit falls (native
59.4 EM / 1.2\% capped; reuse 56.4 / 6.6\%), and the penalty's point estimate moves from $-4.6$ to $-3.0$ (CI
now including zero). But the \emph{direct} paired contrast of the two budgets is $+1.6$ EM (CI $[-0.8,
+4.0]$), which includes zero: we cannot conclude the budget significantly changed the penalty. The higher
cap-hit is evidence of changed generation behavior; its causal contribution to the accuracy gap is not
established.

\paragraph{Replication across adapter initialization} (second seed, same held-out examples). The second-seed
adapter reached a comparable native baseline (53.8 vs.\ 54.4 EM)---a valid replication---with a reuse penalty
of $-0.8$ (CI $[-5.0, +3.2]$). The direct paired seed contrast (seed 2 $-$ seed 1, both 160 tokens) is $+3.8$
EM (CI $[-0.4, +8.2]$), including zero, so the two checkpoints' penalties are not shown to differ.
\textbf{In summary:} across two adapter initialization seeds on the same held-out examples, full-prefix reuse
produced GSM8K accuracy differences of $-4.6$ and $-0.8$ EM at a 160-token budget; raising the first seed's
budget to 320 tokens reduced its observed gap to $-3.0$. All point estimates favored native inference, but
only the first configuration's interval excluded zero. These results neither establish a consistent penalty
magnitude nor demonstrate quality equivalence. Two seeds cannot characterize seed variability; the base-only
gap (${\approx}-45$ EM, under the tested decoding budget) does reproduce, so adapter necessity is robust even
where the reuse penalty is not.

\paragraph{Held-out vs.\ overlapping evaluation.} The GSM8K penalty above ($-4.6$, \texttt{test[580:1080]})
comes from examples untouched by any earlier run. The boundary study below used \texttt{test[80:580]}, which
overlaps prior development, and gave $-1.0\;[-5.0,+3.0]$ for the same full-prefix condition. The two
intervals overlap, so we do \emph{not} claim overlap \emph{caused} the difference; we take the untouched
evaluation to establish a penalty under this protocol, and treat the overlapping one as insufficient
confirmation.

\paragraph{Extractive QA with supplied context.} This result is \emph{not} the full distractor-retrieval
task: for each HotpotQA~\cite{hotpotqa} example we concatenate the distractor context (all 10 paragraphs, 2
gold + 8 distractor) and truncate to 700 tokens, then reuse the base prefix cache of that context for the QA
specialist. Truncation can drop answer-bearing text (not audited), so scores are a lower bound on the
oracle-context setting. At this \emph{short} (700-tok) context, reuse matched native: $\Delta = +0.3$ F1, CI
$[-1.5, +2.2]$ (Table~\ref{tab-qa}). Read with Table~\ref{tab-quality}, the QA reuse penalty is
\textbf{context-dependent}: $+0.3$ at 700 tok, then $-6.6$ (2K) and $-4.5$ (8K)---reuse is not free once the
shared context is long.

\begin{table}[h]\centering
\begin{tabular}{lrr}
\toprule
\textbf{QA specialist (HotpotQA, supplied context, $n=500$)} & \textbf{F1} & \textbf{$\Delta$ vs native} \\
\midrule
Native (specialist prefills context) & 53.0 & --- \\
Full-prefix base-KV reuse & 53.3 & $+0.3\;[-1.5, +2.2]$ \\
aLoRA adapter, native (preliminary) & 42.6 & --- \\
\bottomrule
\end{tabular}
\caption{QA F1 on the supplied-context setting. Native and reuse are the standard-LoRA specialist. The aLoRA
row is a \emph{preliminary} run with matched rank/targets; activation correctness was not verified
(\S\ref{sec-comparisons}), and it is not established to be on the identical sample as the standard-LoRA rows.
The 53.0 here is QA F1 and is unrelated to the numerically-coincident GSM8K native EM of 53.0.}
\label{tab-qa}
\end{table}

\begin{figure}[h]\centering
\includegraphics[width=0.7\textwidth]{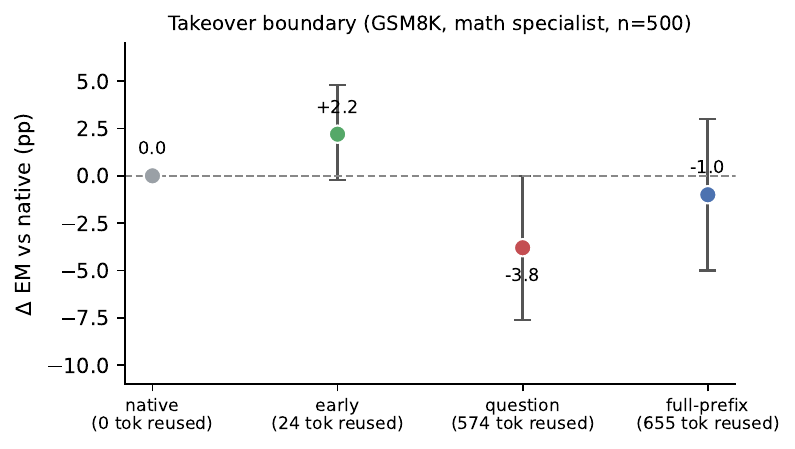}
\caption{$\Delta$ EM vs native as a function of reused-prefix size (paired bootstrap 95\% CIs).
Non-monotonic: reusing more is not uniformly worse. Full-prefix reuse (rightmost) returns closest to
native.}
\label{fig-boundary}
\end{figure}

\paragraph{Takeover boundary (GSM8K, math specialist, $n=500$, overlapping sample \texttt{test[80:580]}).} As
a mechanism probe, moving only $b$ (Table~\ref{tab-boundary}, Figure~\ref{fig-boundary}) gives a non-monotonic
pattern: recomputing more of the prefix is not uniformly better. (This is the overlapping-sample study; the
held-out penalty is above.)

\begin{table}[h]\centering
\begin{tabular}{lrrr}
\toprule
\textbf{boundary} & \textbf{specialist processes} & \textbf{EM} & \textbf{$\Delta$ vs native (paired)} \\
\midrule
Native reference & entire prompt & 53.0 & --- \\
Early takeover & demonstrations + question & 55.2 & $+2.2\;[-0.2, +4.8]$ \\
Question takeover & question only & 49.2 & $-3.8\;[-7.6, +0.0]$ \\
Full-prefix reuse & final prompt token only & 52.0 & $-1.0\;[-5.0, +3.0]$ \\
\bottomrule
\end{tabular}
\caption{Boundary sweep, math specialist, GSM8K, $n=500$ (\texttt{test[80:580]}), paired bootstrap CIs.
Full-prefix reuse is closest to native and cheapest; the mid-context ``question takeover'' has the worst
point estimate but its CI reaches zero.}
\label{tab-boundary}
\end{table}

Two comparisons matter. \textbf{Full-prefix reuse} is closest to native ($\Delta = -1.0$pp, CI $[-5.0,
+3.0]$) and processes the fewest specialist tokens. \textbf{Question takeover}---the intuitive ``share the
background, let the specialist encode the question'' policy---has the worst point estimate ($\Delta =
-3.8$pp, CI $[-7.6, +0.0]$); its direct paired contrast against full-prefix reuse is $-2.8$pp (CI $[-6.6,
+1.0]$), which includes zero.

\paragraph{Interpretation.} In our tested setup full-prefix reuse is cheaper than every other condition and
has a better observed point estimate than question takeover (early takeover has the highest point estimate,
but its interval also includes zero). We do \emph{not} establish quality equivalence (the interval permits a
loss up to ${\approx}5$pp) nor a general boundary-selection rule: \S\ref{sec-quality} does not support
``always recompute more'' or ``never split representations.'' The consistent direction (the mid-context
boundary is the worst cell in every run we did) is a \emph{tendency} we cannot yet attribute to a mechanism.

\section{Comparisons and negative results}\label{sec-comparisons}
\paragraph{Standard LoRA vs Activated LoRA.} Because aLoRA~\cite{alora} is designed for exact base-cache
reuse (adapter weights activate only after an invocation sequence) and its serving engine~\cite{aloraserving}
implements this, it is the natural comparison. We trained an aLoRA adapter on the same QA data (invocation
``\texttt{Answer:}'', matched rank/target modules; Appendix~\ref{app-repro}). This is a \textbf{preliminary
aLoRA run; activation correctness unverified}. Its native F1 was 42.6 versus the standard-LoRA specialist's
53.0, but because we did not validate the activation gating (below), we do \emph{not} attribute this gap to
training quality, capacity, or architecture---it is not yet interpretable. Our structural cached-vs-uncached
parity check for the aLoRA adapter was \textbf{inconclusive}: a plain forward pass does not exercise the
generation-time activation gating that makes aLoRA's reuse exact, so we could not confirm end-to-end
exactness in our harness. A proper comparison (matched standard-LoRA and aLoRA quality, their training
configs, and aLoRA cached-vs-uncached parity under generation) is left as needed work.

\paragraph{A ridge KV translator does not earn its cost.} Motivated by cross-model
transfer~\cite{crossmodelkv,cachebridge}, we fit closed-form per-head ridge maps (RoPE-stripped keys,
single-layer $l{\to}l$ and top-$k$ multi-layer) to translate base KV into the specialist's space. On our
same-backbone setting it did not beat direct reuse on task quality while adding per-layer matrix-multiply
cost per reused token. (For the harder \emph{cross-size} base$\to$base case it improved fidelity---multi-layer
maps reached 78.5\% next-token top-1 agreement---but task retention was 33--56\% and the accurate map was not
economical.) The tested ridge map is rejected; translation as a class is not.

\paragraph{Specialist dependence is not established.} Is the reasoning specialist \emph{specifically} fragile
under reuse? Two contrasts test this and both include zero (Appendix~\ref{app-forest}). The
base-reuse-vs-native penalty difference between adapters was $-9.2$pp (CI $[-21.7, +2.5]$, $n=120$). The
takeover-seam difference-in-differences, $(\text{question}-\text{full-prefix})_{\text{math}} -
(\text{question}-\text{full-prefix})_{\text{QA}}$, was $+2.6$pp (CI $[-3.6, +8.6]$, $n=500$)---and the point
estimate \emph{reverses}: the math seam ($-2.8$pp, CI $[-6.6, +1.0]$) was smaller than the QA seam
($-5.4$pp, CI $[-10.6, -0.2]$). Their individual intervals exclude zero for the math reuse penalty and the QA
seam, but the \emph{differences} between specialists do not. Notably, for the weak-on-task QA specialist,
letting the base encode the question was associated with higher EM (full-prefix 44.2 vs.\ question-takeover
38.8). The present comparisons do not establish specialist dependence; we do not inflate the sample to seek
significance.

\section{Limitations and future work}\label{sec-limits}
The central claim is a quality--latency tradeoff, not equivalence: the GSM8K held-out reuse penalty ($-4.6$)
excludes zero and the QA penalty grows with context. Several gaps bound the claims:

\begin{itemize}
\item \textbf{The penalty is partly truncation.} Reuse doubled the GSM8K generation-cap rate
  (15\%$\to$30\%); a pre-frozen 320-token diagnostic shrinks the penalty from $-4.6$ to $-3.0$ (CI now
  includes zero). Part of the loss is decoding-budget truncation; a residual negative point estimate remains,
  so a budget-independent penalty is neither established nor excluded.
\item \textbf{Two ``shared context'' workloads differ.} Full-prefix reuse as measured shares an
  \emph{identical full prompt} across specialists. Sharing \emph{background passages} across
  \emph{different} questions is a different setting our full-prefix result does not establish; it corresponds
  to a mid-prompt boundary, which is exactly where we see the (uncertain) penalty.
\item \textbf{Provenance.} The $n=500$ boundary evaluation (\texttt{test[80:580]}) is a \emph{larger
  follow-up evaluation}, not an independent held-out confirmation: it overlaps the boundary-development slice
  (\texttt{test[80:200]}) entirely and the earliest development run used \texttt{test[0:500]}. A clean result
  needs a frozen harness and primary comparison, then a fresh evaluation whose size is set for a
  prespecified precision or noninferiority margin.
\item \textbf{Generality.} One backbone, small (1.7B) scale, two tasks. A second-seed replication is done
  (above); its point estimate ($-0.8$) differs from seed 1's ($-4.6$) but the direct contrast includes zero,
  so the penalty magnitude is \emph{uncharacterized}, not shown unstable. Two seeds cannot estimate seed
  variability; replication on a \textbf{different backbone} and more seeds is the remaining generality test,
  alongside re-checking the null specialist-dependence result under those conditions.
\end{itemize}

The boundary sweep is parked; the exploratory measurement history is in Appendix~\ref{app-history}.

\section{Related work}
\paragraph{Cross-model KV transfer.} Heo et al.~\cite{crossmodelkv} fit a closed-form linear mapping to
transfer KV between \emph{different-size} models in a family, reporting 2.7--25$\times$ mapper speedups and
73--98\% accuracy retention on four of six pairs. CacheBridge~\cite{cachebridge} is a concurrent method for
cross-model KV transfer with its own comparisons and results (we do not restate its numbers). Both target the
cross-size case where key/value spaces differ and a map is required. Our setting is the complementary
one---same backbone, different LoRA head---where, for full-prefix reuse, no map is applied; our ridge-map
result is a negative for this regime.

\paragraph{Adapter-aware caching.} Activated LoRA~\cite{alora} modifies the adapter so base-prefix KV is
exactly reusable, and a serving engine~\cite{aloraserving} implements multi-adapter serving on this reuse in
vLLM. We instead ask what already-trained \emph{standard} LoRA adapters preserve under direct reuse, and
quantify it with paired intervals; \S\ref{sec-comparisons} gives our (partial) aLoRA comparison.

\paragraph{Prefix caching.} PagedAttention~\cite{pagedattention} and RadixAttention~\cite{sglang} reuse KV
across requests sharing an \emph{identical} prefix under the \emph{same} model. Our question is reuse of one
prefill across \emph{different} specialists of a shared backbone.

\section{Conclusion}
For composable serving on a shared backbone, reusing the backbone's prefill KV cache across already-trained
standard LoRA specialists is a \textbf{quality--latency tradeoff}. The measured benefit is warm-cache
time-to-first-token, which grows with context (${\approx}16\times$ at 8K); the cost is a small quality loss
whose magnitude was not consistent on GSM8K (held-out $-4.6$ EM at 160 tokens, $-3.0$ at 320, $-0.8$ under a
second seed---all favoring native, only the first excluding zero) and context-dependent on QA ($-6.6$/$-4.5$
F1 at 2K/8K). The memory story is the paper's main correction: this implementation reuses KV \emph{values}
but \emph{copies their storage}---two-branch peak was only 12\% lower at 8K and the prefix was never
physically shared, so shared-cache memory savings are \emph{not} achieved and would need a paged cache.
Partial recomputation gave no demonstrated advantage; a ridge translator did not earn its cost;
specialist-dependence did not resolve. The immediate next step is the frozen generation-budget diagnostic
(to settle the truncation question), then a replication on another seed/backbone; the claims above should be
confirmed on one's own adapters before reuse is assumed lossless.

\clearpage
\appendix
\section{Specialist-dependence contrasts}\label{app-forest}
\begin{figure}[h]\centering
\includegraphics[width=0.86\textwidth]{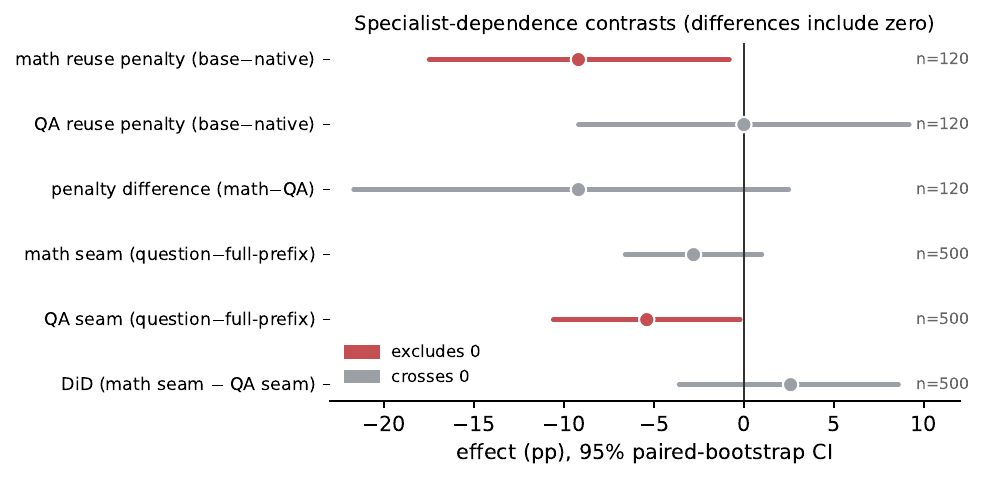}
\caption{Contrasts bearing on specialist-specific reuse penalty. Red intervals exclude zero; grey include
it. The individual math reuse penalty and QA seam exclude zero, but the \emph{differences} between
specialists---the penalty difference ($n=120$) and the seam difference-in-differences ($n=500$)---include
zero, and the DiD point estimate is positive. Note the differing $n$; these come from separate runs.}
\label{fig-forest}
\end{figure}

\section{Exploratory measurement history}\label{app-history}
The ``question takeover'' penalty changed across successive \emph{development} measurements
(Table~\ref{tab-drift}). These are not independent replications: sample draw and prompt construction changed
together. One difference we can rule out is \texttt{add\_special\_tokens} (a no-op here---the Qwen3 tokenizer
emits no BOS token, verified); the rest (a shared instruction preamble; different, overlapping GSM8K slices)
are confounded. We report the final row and treat the earlier ones as development history, not evidence of a
fixed bug.

\begin{table}[h]\centering
\begin{tabular}{lrlll}
\toprule
\textbf{run (role)} & \textbf{n} & \textbf{prefix} & \textbf{test slice} & \textbf{$\Delta$ (question)} \\
\midrule
initial (dev) & 500 & demos & \texttt{test[0:500]} & $-14.4$ \\
matrix (dev) & 120 & demos & \texttt{test[0:120]} & $-9.2$ \\
boundary (dev) & 120 & instr+demos & \texttt{test[80:200]} & $-4.2$ \\
confirm (follow-up) & 500 & instr+demos & \texttt{test[80:580]} & $-3.8$ \\
\bottomrule
\end{tabular}
\caption{Development history of the ``question takeover'' penalty. The final follow-up overlaps the boundary
development slice (\texttt{test[80:200]} $\subset$ \texttt{test[80:580]}); it is not an independent held-out
sample.}
\label{tab-drift}
\end{table}

\section{Reproducibility}\label{app-repro}
Code, adapter checkpoints, exact configs, prompts, and per-example score arrays are at
\url{https://github.com/AltSlate-Labs/routekv}. Remaining unpinned items (below) are marked; this appendix is
an outline, and the fully-pinned artifact accompanies the repository.

\paragraph{Models.} Backbone \texttt{Qwen/Qwen3-1.7B} (bf16; HF revision \texttt{main}, exact commit pinned
in the repo). Specialists are LoRA~\cite{lora} adapters, rank 16, $\alpha=32$, dropout 0, target modules
\texttt{q\_proj,k\_proj,v\_proj,o\_proj}; QA trained on HotpotQA~\cite{hotpotqa} (distractor split), math on
GSM8K~\cite{gsm8k} (main), plus one aLoRA~\cite{alora} QA adapter (invocation ``\texttt{Answer:}'', matched
rank/targets). Per-adapter SFT recipe, seeds, and checkpoint hashes are in the repo (training seeds were not
varied---a single seed per task, which is why replication across seeds is future work).

\paragraph{Evaluation.} GSM8K: 4 fixed few-shot demonstrations from the train split; prompt
\texttt{[instruction | demos | "Question: \{q\}$\backslash$nAnswer:"]}; greedy, generation capped at 160
tokens. Answer extraction removes thousands-separators (commas), then takes the last signed-integer match;
\emph{decimals and fractions are not parsed} and a completion with no integer scores as wrong. The fraction
of completions that hit the 160-token cap was \emph{not logged} in these runs (it is reported in the planned
matched experiment, \S\ref{sec-limits}); cap-induced truncation could bias adapter comparisons and is a known
gap. QA: for each example the distractor context (all 10 paragraphs, 2 gold + 8 distractor) is concatenated
and truncated to 700 tokens (truncation can drop answer text; not audited), token-level F1 on the first
output line. Sample ranges: boundary follow-up GSM8K \texttt{test[80:580]} ($n=500$, overlapping earlier
development slices, Appendix~\ref{app-history}); own-KV control and penalty-difference contrast $n=120$; QA
$n=500$. Paired bootstrap CIs use 10,000 resamples of per-example differences~\cite{bootstrap}; per-example
arrays are released.

\paragraph{Systems measurement.} transformers 5.5 / peft 0.20 / torch 2.11, one NVIDIA RTX PRO 4500 Blackwell
GPU; attention backend and exact timing protocol (warmup, repetitions, CUDA synchronization) are pinned in
the repo---reported timings here are single-run. Table~\ref{tab-cost} (single-prefix cost),
Table~\ref{tab-serving} (warm-cache latency), and Table~\ref{tab-residency} (two-branch peak) are separate
runs. Prefix-cache figures in Table~\ref{tab-cost} are KV tensor footprint. The physical-sharing check
(Table~\ref{tab-residency}) is a \texttt{data\_ptr} identity test across the two branches' prefix tensors,
measured before and after generation.

\paragraph{Translator.} Per-head ridge maps calibrated on 40 held-out HotpotQA distractor contexts
(\texttt{N\_CAL=40}) with RoPE-stripped keys; single-layer $l{\to}l$ and top-$k$ multi-layer variants; ridge
regularization $\lambda=10$. Reported cross-size fidelity (next-token top-1 agreement) and task retention are
from that calibration; the numerical comparison against direct reuse (no quality gain, added per-token
matrix-multiply cost) is in the repo. The map was not adopted.

\bibliographystyle{unsrt}
\bibliography{refs}
\end{document}